\documentclass[pdflatex,sn-mathphys-num]{sn-jnl}% Math and Physical Sciences Numbered Reference Style
\usepackage{graphicx}%
\usepackage{multirow}%
\usepackage{amsmath,amssymb,amsfonts}%
\usepackage{amsthm}%
\usepackage{mathrsfs}%
\usepackage[title]{appendix}%
\usepackage{xcolor}%
\usepackage{textcomp}%
\usepackage{manyfoot}%
\usepackage{booktabs}%
\usepackage{algorithm}%
\usepackage{algorithmicx}%
\usepackage{algpseudocode}%
\usepackage{listings}%
\usepackage{makecell}
\usepackage{tikz}
\usetikzlibrary{positioning, shapes, fit, backgrounds, arrows.meta, calc, decorations.pathreplacing,calligraphy, matrix}
\usepackage{pgfplots}
\usepackage{pgfplotstable}
\usepackage{xcolor}
\pgfplotsset{compat=newest}
\usepackage[table]{xcolor} 
\usepackage{colortbl} 
\theoremstyle{thmstyleone}%
\theoremstyle{thmstyletwo}%

\theoremstyle{thmstylethree}%

\begin{document}

\title[ECG-LLM]{ECG-LLM-- training and evaluation of domain-specific large language models for electrocardiography }

%%=============================================================%%
%% GivenName	-> \fnm{Joergen W.}
%% Particle	-> \spfx{van der} -> surname prefix
%% FamilyName	-> \sur{Ploeg}
%% Suffix	-> \sfx{IV}
%% \author*[1,2]{\fnm{Joergen W.} \spfx{van der} \sur{Ploeg} 
%%  \sfx{IV}}\email{iauthor@gmail.com}
%%=============================================================%%

\author[1]{\fnm{Lara} \sur{Ahrens}}\email{lara.ahrens@uol.de}

\author[2]{\fnm{Wilhelm} \sur{Haverkamp}}\email{wilhelm.haverkamp@dhzc.charite.de}

\author*[1]{\fnm{Nils} \sur{Strodthoff}}\email{nils.strodthoff@uol.de}

\affil[1]{\orgdiv{AI4Health Division}, \orgname{Carl von Ossietzky Universität Oldenburg}, \orgaddress{\city{Oldenburg}, \country{Germany}}}

\affil[2]{\orgdiv{Department of Cardiology, Angiology and Intensive Care Medicine}, \orgname{Charité Campus Mitte, German Heart Center of the Charité-University Medicine Berlin}, \orgaddress{\city{Berlin}, \country{Germany}}}

%%==================================%%
%% Sample for unstructured abstract %%
%%==================================%%

\abstract{
Domain-adapted open-weight large language models (LLMs) offer promising healthcare applications, from queryable knowledge bases to multimodal assistants, with the crucial advantage of local deployment for privacy preservation. However, optimal adaptation strategies, evaluation methodologies, and performance relative to general-purpose LLMs remain poorly characterized.
We investigated these questions in electrocardiography, an important area of cardiovascular medicine, by finetuning open-weight models on domain-specific literature and implementing a multi-layered evaluation framework comparing finetuned models, retrieval-augmented generation (RAG), and Claude Sonnet 3.7 as a representative general-purpose model.
Finetuned Llama 3.1 70B achieved superior performance on multiple-choice evaluations and automatic text metrics, ranking second to Claude 3.7 in LLM-as-a-judge assessments. Human expert evaluation favored Claude 3.7 and RAG approaches for complex queries. Finetuned models significantly outperformed their base counterparts across nearly all evaluation modes.
Our findings reveal substantial performance heterogeneity across evaluation methodologies, underscoring assessment complexity. Nevertheless, domain-specific adaptation through finetuning and RAG achieves competitive performance with proprietary models, supporting the viability of privacy-preserving, locally deployable clinical solutions.
}

\keywords{Natural Language Processing, Large Language Models, Domain Specialization, Cardiology, Electrocardiography}

%%\pacs[JEL Classification]{D8, H51}

%%\pacs[MSC Classification]{35A01, 65L10, 65L12, 65L20, 65L70}

\maketitle

\section{Introduction}\label{sec:intro}
\begin{figure}[htbp] % htbp = here, top, bottom, page
  \centering
  \includegraphics[width=\textwidth]{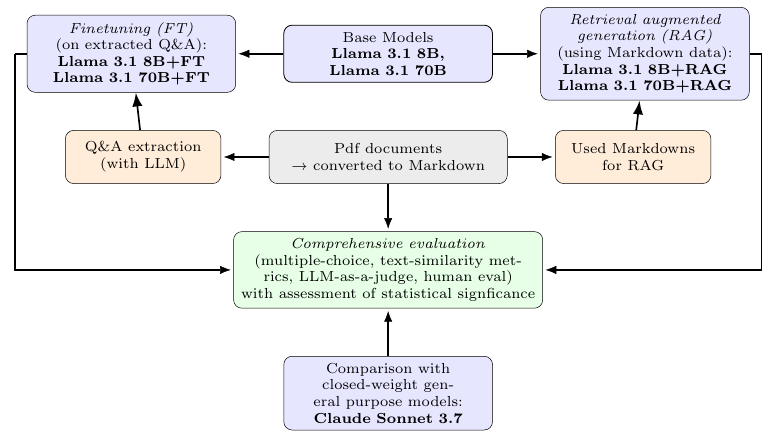}
  \caption{Schematic overview of the core methodology of this study: We study finetuning and retrieval augmented generation as two possible paths towards domain specialization of Llama 3 open-weight LLMs in the domain of electrocardiography. Model performance is assessed in a comprehensive evaluation procedure covering four different categories comparing domain-specialized models in comparison to the respective base models and a commercial general-purpose LLM.} 
  \label{fig:methods} 
\end{figure}

Large-language models (LLMs) used as question-answering models \cite{kell2024question} represent a promising approach for clinicians to cope with the ever-increasing amount of medical knowledge. If properly validated and implemented, such systems could help reduce misdiagnoses and associated treatment errors and could support the identification of optimal, evidence-based treatment strategies. Such a paradigm could be implemented either through commercial general-purpose LLMs, with associated privacy risks, or through smaller domain-adapted open-weight models.
In this study, we focus on cardiology with an emphasis on electrocardiography and set out to detail practical techniques for domain specialization of LLMs and compare them thoroughly in a comprehensive evaluation approach; see Figure~\ref{fig:methods} for a schematic overview.

%\heading{Rise of Large Language Models}
For natural language processing (NLP) tasks such as question answering, large language models (LLMs) are becoming increasingly important. Models with varying capabilities, sizes, and knowledge bases are currently being developed \cite{10.1145/3744746}. However, LLMs tend to hallucinate when knowledge, in particular domain-specific knowledge, is missing, generating incorrect or fabricated information \cite{huang2023survey}, which represents a major hurdle in safety-critical domains such as medicine. The most powerful LLMs, such as GPT-4o and Claude Sonnet 3.7, are typically closed-source models, which prohibits local hosting and limits domain knowledge adaptation. In the medical domain, additional challenges include specialized knowledge, specialized vocabulary, reliability, and data protection \cite{li-etal-2024-llamacare}. Consequently, large closed-source models may not be ideal for medical tasks. Therefore, we explore smaller open-weight LLMs that can be locally hosted and adapted to specific medical contexts.

%\heading{Domain-Specific Language}
However, the superiority of domain-specialized language models over their general-purpose counterparts should not be taken for granted. Prior work \cite{jeong2024medical} found no evidence for significant improvements when using appropriate prompting, and the same holds for finetuned commercial LLMs \cite{Wu2025}. Nevertheless, finetuned domain-specialized models have been explored in many domains, including medical subdomains such as radiology \cite{ranjit2024radphi3smalllanguagemodels}, material science \cite{schillingwilhelmi2024textinsightlargelanguage}, and mathematics \cite{toshniwal2024openmathinstruct2acceleratingaimath}.

%\heading{Methods for Domain Specialization}
Identifying the most effective methods for domain specialization of pretrained general-purpose LLMs remains an active area of research. Possible methods range from continual pretraining \cite{yang2024syntheticcontinuedpretraining} and supervised finetuning \cite{zhao2025stylefactsmappingboundaries,wu2024finetunebenchcommercialfinetuningapis,lin2025sft} to retrieval-augmented generation (RAG) \cite{Liu.03.10.2024,Kamath.2024,ong2024surgeryllm}. Whereas finetuning was previously believed to primarily adjust output style, recent work has demonstrated its potential for knowledge injection through question-answer pairs \cite{zhao2025stylefactsmappingboundaries}. This aligns with previous research showing that finetuning can fundamentally improve the model's domain knowledge \cite{li-etal-2024-llamacare, Haghighi., yang2023investlmlargelanguagemodel, 10704843, Chao2025}. RAG represents a complementary approach that enriches the model's responses with external, factual information \cite{Liu.03.10.2024}. During RAG, the LLM is supplemented with retrieved, contextually relevant data to generate fact-based and verifiable answers \cite{Kamath.2024}.

%\heading{LLMs for Cardiology}
Comprehensive reviews highlight the growing potential of LLMs in cardiovascular medicine, including applications in patient education, clinical decision support, and workflow automation \cite{Santos2025LLMCardiologyReview}. Studies have demonstrated that finetuning LLMs on small-scale cardiology datasets can substantially improve clinical text classification and information extraction \cite{losch2025finetuningllmssmallmedical}, and LLMs have shown promise in automating discharge summary generation for cardiac patients \cite{jung2024enhancingclinicalefficiencyllm}. In evaluation studies, Lee et al.\ benchmarked general-purpose LLMs on cardiology case questions modeled after the American College of Physicians' MKSAP examination, finding that GPT-4 achieved performance comparable to seasoned cardiologists \cite{Lee2023LLMCardiology}. While these studies demonstrate the potential of LLMs in cardiology, they either focus on narrow application domains or do not investigate the effects of domain specialization through comprehensive evaluation. A recent study \cite{Chao2025} demonstrated that a finetuned Llama 2 outperformed its original base model (the starting point before domain adaptation) across all automated metrics and human evaluations. Our work builds on state-of-the-art base models from the Llama 3 family \cite{dubey2024llama3herdmodels} and provides a more comprehensive evaluation. Regarding RAG, a recent cardiology study showed that RAG can improve LLMs' knowledge of ECG and outperform OpenAI models across different tasks \cite{ragllmecg}. This RAG evaluation is closely related to our work, which additionally provides a comparative assessment of finetuning and a more comprehensive evaluation comprising different methods.

%\heading{Towards multimodal assistant models}
Recent works have extended the scope to multimodal applications that leverage LLMs as a core component, including multimodal multi-agent models \cite{zhou2024zodiaccardiologistlevelllmframework,zhang2025cardaicagentsmultimodalframeworkhierarchical}, conversational ECG interpretation models \cite{zhao2025ecgchatlargeecglanguagemodel}, and multimodal medical text retrieval models \cite{pham2025qheartecgquestionanswering}. Since most multimodal assistant models leverage general-purpose LLMs, they could potentially benefit from domain-adapted LLMs as explored in this work.

\section{Results}\label{sec:results}
We finetune Llama 3.1 8B and 70B open-weight (instruct) models on question-answer pairs extracted from domain-specific literature by prompting Llama 3.3 70B. We use the same references as a basis for retrieval-augmented generation and evaluate models using multiple-choice questions extracted from the same corpus, automatic text similarity metrics, LLM-as-a-judge evaluation, and human expert evaluation by an experienced cardiologist. We employ empirical bootstrapping to derive statistically robust rankings.

\subsection{Multiple-choice evaluation} For the multiple-choice evaluation, we assess three dataset subsets: ``full'' contains the entire dataset excluding training data (27,774 samples), ``special'' contains questions from documents highlighted by the human expert as particularly relevant (1,219 samples), and ``checked'' contains multiple-choice questions verified by a human expert with long-standing experience in electrocardiography and cardiac electrophysiology (534 samples). Table~\ref{tab:acc_results_full} presents the results across these three tests. The finetuned versions and RAG-enhanced models consistently outperform general-purpose models, with the two domain-specialized 70B models performing best. Notably, although model accuracies vary slightly, the ranking remains consistent across all three subsets.

\begin{table}[htbp]
\centering
\caption{Performance of different models across multiple-choice evaluation sets. Values represent accuracy percentages with rank in parentheses. Domain-adapted models explored in this work are highlighted in bold face. Domain-adapted models dominate over general-purpose models in this setting. Special = specialized cardiovascular/ECG questions; Full = complete question set; Checked = manually validated subset; RAG = retrieval-augmented generation; FT = finetuning. }
\label{tab:acc_results_full}
\begin{tabular}{@{}lrrr@{}}
\toprule
\textbf{Model} & \textbf{Special} & \textbf{Full} & \textbf{Checked} \\
\midrule
\textbf{Llama 3.1 70B + FT} & 90.2 (1) & 92.0 (1) & 88.2 (1) \\
\textbf{Llama 3.1 70B + RAG} & 87.9 (2) & 90.2 (2) & 86.5 (2) \\
\textbf{Llama 3.1 8B + RAG} & 86.1 (3) & 88.5 (3) & 85.0 (3) \\
\textbf{Llama 3.1 8B + FT} & 84.9 (3) & 87.5 (6) & 84.3 (4) \\
Claude Sonnet 3.7 & 82.3 (5) & 88.0 (4) & 81.7 (5) \\
Llama 3.1 70B & 82.4 (6) & 88.2 (5) & 81.5 (6) \\
Llama 3.1 8B& 73.3 (7) & 81.6 (7) & 73.0 (7) \\
\bottomrule
\end{tabular}
\end{table}
\subsection{Text similarity metrics}
For automatic evaluation, we assessed free text responses using BLEU \cite{bleu_score}, ROUGE-1, ROUGE-2, ROUGE-L \cite{lin2004rouge}, and BERTScore \cite{bertscore}, where BLEU and ROUGE assess surface similarity and BERTScore assesses semantic similarity between generated response and the reference answer. Table~\ref{tab:bleu_rouge1} shows the F1 scores for these metrics; recall and precision values are provided in the Appendix \ref{ap:otherTestedConfig}. The finetuned models consistently outperform state-of-the-art models, while RAG-enhanced models rank last, performing worse than the base models. Again, the ranking remains consistent across the different metrics, which assess different levels of text similarity.

\begin{table}[htbp]
\centering
\caption{Text summarization performance metrics across different models. Values represent F1 scores (ROUGE, BERTScore) or BLEU scores with rank in parentheses. Domain-adapted models explored in this work are highlighted in bold face. Finetuned models excel in this category. RAG = retrieval-augmented generation; FT = finetuning. }
\label{tab:bleu_rouge1}
\begin{tabular}{@{}lrrrrr@{}}
\toprule
\textbf{Model} & \textbf{ROUGE-1} & \textbf{ROUGE-2} & \textbf{ROUGE-L} & \textbf{BLEU} & \textbf{BERTScore} \\
 & \textbf{F1} & \textbf{F1} & \textbf{F1} & & \textbf{F1} \\
\midrule
\textbf{Llama 3.1 70B + FT} & 0.4270 (1) & 0.2449 (1) & 0.3764 (1) & 0.1289 (1) & 0.3904 (1) \\
\textbf{Llama 3.1 8B + FT} & 0.4063 (2) & 0.2279 (2) & 0.3568 (2) & 0.1195 (2) & 0.3700 (2) \\
Llama 3.1 70B& 0.3852 (3) & 0.1981 (3) & 0.3208 (3) & 0.0962 (3) & 0.3476 (3) \\
Claude Sonnet 3.7 & 0.3661 (4) & 0.1685 (5) & 0.2992 (4) & 0.0694 (6) & 0.3395 (4) \\
Llama 3.1 8B & 0.3537 (5) & 0.1730 (4) & 0.2906 (5) & 0.0788 (4) & 0.2936 (5) \\
\textbf{Llama 3.1 8B + RAG} & 0.3481 (6) & 0.1689 (5) & 0.2831 (6) & 0.0763 (5) & 0.2861 (6) \\
\textbf{Llama 3.1 70B + RAG} & 0.2557 (7) & 0.1490 (7) & 0.2183 (7) & 0.0622 (7) & 0.0181 (7) \\
\bottomrule
\end{tabular}
\end{table}

\subsection{LLM-as-a-judge}
\begin{figure}[htbp]
    \centering
    \includegraphics[]{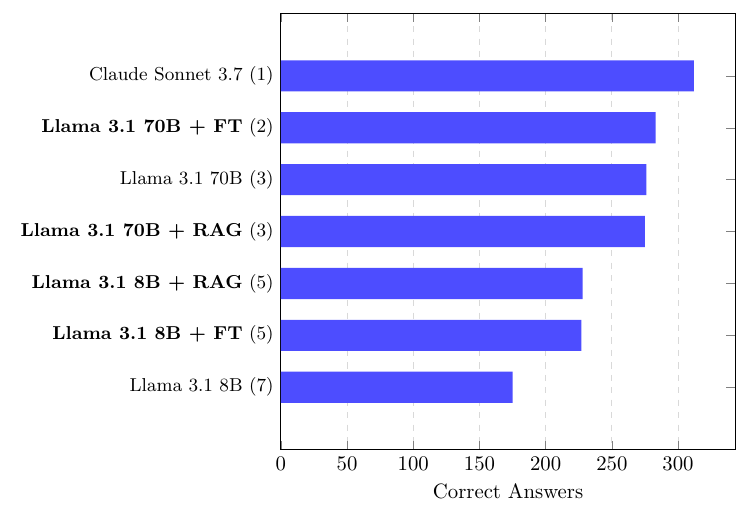}
    \caption{Correct answers resulting from 417 questions evaluated with LLM-as-a-judge. The number in parentheses after the model name indicates the rank in the statistically robust model ranking. Domain-adapted models explored in this work are highlighted in bold face. Claude Sonnet 3.7 performs best, closely followed by Llama 3.1 70B models. RAG = retrieval-augmented generation; FT = finetuning.}
    \label{fig:llmJudge}
\end{figure}

For the LLM-as-a-judge evaluation, a subset of the question-and-answer dataset was checked manually and corrected by a human expert to ensure quality. From this subset, we selected questions not part of the training dataset, resulting in 417 questions for evaluation. Figure~\ref{fig:llmJudge} shows the results. Claude achieves the best results, while the finetuned 8B model shows substantial improvement, answering approximately 50 more questions correctly than its base model Llama 3.1 8B. The finetuned 70B model also improves, while other models remain stable or decrease in performance. Llama 3.1 8B and 70B with RAG achieve results similar to the finetuned versions. This metric shows statistically significant improvements in the finetuned models compared to their respective base models.

To verify the reliability of the LLM-as-a-judge assessment, a human expert evaluated a random subset of answers from the finetuned models labeled as correct or incorrect. The results are compiled in Appendix~\ref{app:llmaaj}. Overall, 80\% of evaluations showed agreement between the LLM and human expert. Notably, all answers judged correct by the LLM were confirmed correct, but nearly 50\% of answers judged incorrect by the LLM were considered correct by the human expert.

\subsection{Human evaluation}
For the human evaluation, the expert provided ten questions asking for facts similar to those used for finetuning, as well as 40 more complex questions investigating relationships and differing in format and context. These complex questions, for example, ask for listings or descriptions of relationships between different characteristics. The expert then labeled the answers as incorrect, partially incorrect, correct but incomplete, or correct. The questions, model responses and expert evaluations are provided in the associated code repository.

\begin{figure}[ht]
    \centering
    \includegraphics[]{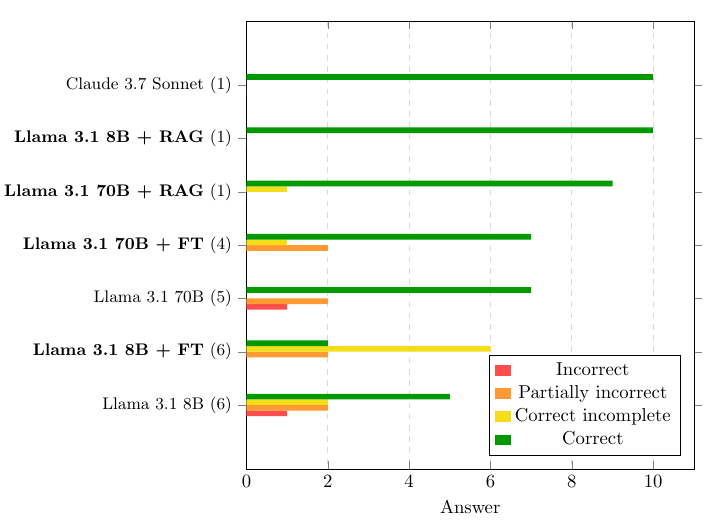}
    \caption{Human evaluation of the ten factual questions. The number in parentheses after the model name indicates the rank in the statistically robust model ranking.  Domain-adapted models explored in this work are highlighted in bold face. Both RAG models perform on par with the top-performing Claude 3.7 Sonnet. RAG = retrieval-augmented generation; FT = finetuning.}
    \label{fig:human_10}
\end{figure}

%\heading{Factual questions} 
Figure~\ref{fig:human_10} shows the results of the human evaluation for the ten factual questions. The finetuned versions perform similarly to or better than their corresponding base models, providing more correct and correct-but-incomplete answers in total and never answering questions completely incorrectly. However, no model outperformed Claude Sonnet 3.7 and Llama 3.1 8B with RAG, which both answered all questions correctly. Llama 3.1 70B with RAG achieved one incomplete answer. 

\begin{figure}[ht]
    \centering
    \includegraphics[]{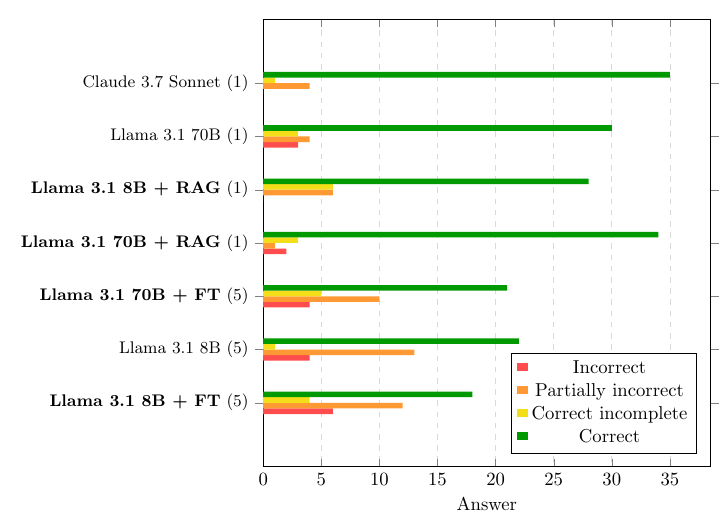}
    \caption{Human evaluation of the 40 semantically complex questions. The number in parentheses after the model name indicates the rank in the statistically robust model ranking. Both RAG models as well as the Llama 3.1 70B base model perform on par with the best-performing Claude 3.7 Sonnet. Domain-adapted models explored in this work are highlighted in bold face. RAG = retrieval-augmented generation; FT = finetuning.}
    \label{fig:human_40}
\end{figure}

%\heading{More complex questions}
Figure~\ref{fig:human_40} shows the results for the semantically more complex questions. Notably, the finetuned versions demonstrate slightly decreased performance compared to their base models. The number of completely incorrect answers increases for each finetuned model, resulting in rankings similar to Llama 3.1 8B. In contrast, Llama 3.1 8B with RAG outperforms its base model and reaches the performance of Llama 3.1 70B. Unlike other evaluations, the 40-question assessment shows finetuned models either underperforming or achieving similar results to their corresponding base models.

%\heading{Insights from text complexity} 
To analyze the differences between the training data and the 40 complex human-provided questions, we calculated the average Flesch reading ease score. This metric measures text difficulty by analyzing word-to-sentence and syllable-to-word ratios, yielding a score between 0 and 100, where lower values indicate higher difficulty \cite{MohammedLubnaAli2022AEfa}. The human-provided questions averaged a Flesch score of 21.9, while the training data averaged 30.3, confirming that the human questions are semantically more complex.

\subsection{Overall ranking}
Table~\ref{tab:median_ranks} shows the median ranks of each model across all evaluation methods. The finetuned Llama 3.1 70B demonstrates the strongest overall performance, followed by Claude Sonnet 3.7, which underperforms in multiple-choice and text-similarity evaluations. RAG is effective in some categories (LLM-as-a-judge and human evaluation) but underperforms in others and fails to consistently outperform the corresponding base models. Larger models (70B) consistently outperform smaller models (8B).

\begin{table}[htbp]
\centering
\caption{Ranks for each of the models across the different evaluation categories (reporting medians in case of multiple results per category). Models are listed according to median rank across all categories. RAG = retrieval-augmented generation; FT = finetuning. }
\label{tab:median_ranks}
\begin{tabular}{@{}lrrrrr@{}}
\toprule
\textbf{Model} & \textbf{Multiple-} & \textbf{Text Similarity} & \textbf{LLM-as-a-} & \textbf{Human} & \textbf{Median} \\
 & \textbf{Choice} & \textbf{Metrics} & \textbf{Judge} & \textbf{Evaluation} & \textbf{Rank} \\
\midrule
\textbf{Llama 3.1 70B + FT} & 1 & 1 & 2 & 3 & 1.5 \\
Claude Sonnet 3.7 & 5 & 4 & 1 &1 & 2.5 \\
\textbf{Llama 3.1 70B + RAG} & 3 & 7 & 3 &1&3 \\
Llama 3.1 70B& 6&3 &3 & 3 & 3 \\
\textbf{Llama 3.1 8B + RAG} & 2 & 6 & 5 &1&3.5  \\
\textbf{Llama 3.1 8B + FT} & 4 & 2 & 5& 5.5 & 4.5\\
Llama 3.1 8B & 7 & 5 & 7 & 5.5 & 6\\
\bottomrule
\end{tabular}
\end{table}

\section{Discussion}\label{sec:discussion}
\subsection{Evaluation modes}
%\heading{Multiple-choice evaluation} 
Because the multiple-choice questions are LLM-generated, they may lack variation, and incorrect answer options might be obvious or occasionally correct. Nevertheless, consistent model rankings across all subsets, including the human-checked subset, suggest reliable results. Notably, the finetuned models outperform all state-of-the-art models on the checked and special subsets, indicating successful knowledge injection during finetuning. However, on the full multiple-choice set, the finetuned 8B model underperforms Claude and Llama 3.1 70B. The varying improvement margins also indicate that models with stronger baselines benefit less from finetuning.

%\heading{Text similarity metrics}
Notably, all baseline models show poor performance in automatic text similarity evaluations. Since these metrics assess syntactic and semantic similarity \cite{bleu_score, lin2004rouge, bertscore}, they demonstrate that finetuned responses align with reference answers, confirming knowledge acquisition. However, they cannot prove that state-of-the-art models underperform, as alternative correct formulations would score poorly despite being accurate. While these metrics effectively evaluate stylistic consistency of finetuned responses \cite{zhao2025stylefactsmappingboundaries}, they are less reliable for baseline models using alternative formulations. Evaluation based solely on BLEU and ROUGE scores can reveal style injection but cannot confirm superior knowledge. However, the agreement across all metrics, including BERTScore's higher-level semantic alignment, supports the robustness of these findings.

%\heading{LLM-as-a-judge evaluation}\label{sec:discussionllamajudge}
The LLM-as-a-judge also ranks finetuned models higher than their base versions. However, this evaluation may be influenced by stylistic alignment, favoring answers matching the judge LLM's preferences. Furthermore, the judge LLM's internal judgment mechanisms are unknown, making it unclear whether it prioritizes reasoning or stylistic coherence. Nevertheless, finetuned models achieve more correct answers than their base versions, and RAG models perform similarly, as confirmed by other evaluations. This raises questions about the sharpness of distinctions between correct and incorrect answers; more refined evaluation categories, similar to those in human evaluation, could provide clarity. Since this method correlates well with human evaluations for correct answers and scales efficiently, it is suitable for automated evaluation when sufficient contextual information is available.

%\heading{Human evaluation}
Human evaluation is considered the gold standard but, unlike other quality metrics, cannot be scaled to large sample sizes. Nevertheless, even with 10-40 questions as in this work, it can reveal statistically significant performance differences. Due to the limited sample size, human evaluation typically does not comprehensively cover all relevant subdomains. Additionally, questions from a small group of experts introduce bias in question style. More semantically complex questions, as observed here, may advantage larger models with better general language understanding. Finally, we acknowledge that our evaluation scheme considers only factual correctness and does not involve direct pairwise preference evaluation, such as Elo scores \cite{chiang2024chatbot}. Since the models already differ substantially in factual correctness, we consider pairwise comparisons a promising next step.

To summarize, this work demonstrates the importance of multi-faceted assessment approaches, as highlighted by \cite{Teo2025GenerativeAI}. No single evaluation method provides comprehensive coverage of model capabilities, with each approach offering distinct advantages and limitations. Our evaluation differs from previous studies that typically rely on one or two methods \cite{Haghighi., li-etal-2024-llamacare, 10704843, huang2023survey, Chao2025}, revealing limitations that narrower evaluations may overlook. This concern was also emphasized by \cite{jeong2024medicaladaptationlargelanguage, Teo2025GenerativeAI}.
Statistical significance analysis is crucial to our methodology, as all evaluation methods are influenced by statistical variation. This leads to models achieving similar rankings despite different scores, making it difficult to establish clear performance hierarchies. Consequently, performance differences reported in other studies \cite{Haghighi., li-etal-2024-llamacare, 10704843, Chao2025} may be attributable to chance rather than meaningful distinctions, as cautioned in \cite{jeong2024medicaladaptationlargelanguage}.

\subsection{Methods for domain specialization}

The results demonstrate that finetuning enhances factual knowledge in domain-specific tasks. In in-distribution tests, which consisted of  multiple-choice evaluations and automated text similarity metrics, finetuned models often match or exceed larger state-of-the-art models like Llama 3.1 70B and Claude Sonnet 3.7, aligning with prior work \cite{Haghighi., li-etal-2024-llamacare, 10704843, yang2023investlmlargelanguagemodel, zhao2025stylefactsmappingboundaries, Teo2025GenerativeAI, Chao2025}. This confirms that domain-specific finetuning is effective in specialized areas like ECG without continual pretraining.
However, human evaluation reveals limitations: Finetuned models underperform on questions syntactically different from training data, where smaller state-of-the-art or base models may excel. This likely results from insufficient complexity and diversity in the finetuning dataset \cite{zhao2025stylefactsmappingboundaries}. Solutions include more complex training data or pretraining \cite{Haghighi.}, though one study the effectiveness of the latter \cite{jeong2024medicaladaptationlargelanguage}. Additionally, synthetic data generation via LLM prompting lacks control over diversity and complexity, potentially limiting finetuning results. Human quality assurance \cite{zheng2025miriadaugmentingllmsmillions} and more varied training data could address these issues.
Overall, finetuning enables strong in-distribution performance but struggles with out-of-distribution queries due to limited data diversity, highlighting areas for refinement in synthetic data generation.

%\heading{RAG}
RAG-enhanced models perform similarly to finetuned models across most metrics. On in-distribution data, especially multiple-choice questions, they match finetuned models and can outperform larger models. However, they show notably decreased performance on BLEU, ROUGE, and BERTScore evaluations, supporting that lexical metrics are less meaningful in isolation. On out-of-distribution data in human evaluation, RAG consistently outperforms finetuned models and reaches Claude's performance, aligning with \cite{ragllmecg}. As shown in previous work \cite{ong2024surgeryllm, Teo2025GenerativeAI}, RAG offers a flexible option for knowledge enhancement. Notably, Llama 3.1 8B with RAG matches larger models while enabling local hosting for data control and independence from external dependencies—essential factors in medicine.

%\heading{Large-scale proprietary models}
Claude Sonnet 3.7, representing large-scale proprietary models, is outperformed on multiple-choice, BLEU, ROUGE, and BERTScore evaluations but demonstrates strong performance in human evaluation. This shows that while imperfect on specific details, general-purpose models possess strong broad knowledge also in specialized domains such as ECG analysis. One study demonstrates that improvements in one domain can transfer to others without explicit training \cite{li2025domainhelpothersdatacentric}, which alongside likely extensive finetuning on PubMed data \cite{jeong2024medicaladaptationlargelanguage} may explain the strong performance of large-scale proprietary models.

%\heading{Comparison Finetuning vs. RAG}
Turing to a direct comparison, both RAG and finetuning increase ECG domain knowledge. Finetuning outperforms RAG on specialized data distributions, as demonstrated in automatic evaluations with syntactic and semantic constraints. However, performance remains similar in other in-distribution tests, indicating that further analysis is needed to determine exact differences. Human evaluations suggest base models perform better on out-of-distribution queries.
Production suitability depends on factors such as document accessibility, vector database availability, and intended use. RAG offers a clear advantage: easy updates to reflect evolving literature, whereas finetuning requires costly retraining.

%\heading{Future work}
To improve finetuned models, continual pretraining followed by supervised finetuning shows promise \cite{Haghighi.}, though one study questions whether pretraining yields substantial improvements over state-of-the-art models \cite{jeong2024medicaladaptationlargelanguage}. A more varied training dataset covering diverse medical tasks (diagnoses, recommendations, complex relationships) and including human-generated data or complex chat instructions could significantly improve performance. Another option is finetuning with RLHF, where human feedback trains a reward model to guide the LLM toward defined preferences \cite{rlhffinetuning}. RAG implementation could also be improved through deeper analysis of embedding models, finetuning embeddings, or curating document selection.
Combining RAG and finetuning warrants evaluation to leverage both approaches' advantages. Since finetuned models underperform on syntactically and semantically different questions, further investigation is needed to determine whether they can effectively use contextual information or require different finetuning strategies for RAG integration.

\subsection{Clinical application scenarios}

Our proposed models could enhance ECG interpretation accuracy and speed in clinical ECG practice, supporting immediate decision-making for tasks such as arrhythmia identification, ST-segment analysis, and QT assessment \cite{Santos2025LLMCardiologyReview}. By providing instant, evidence-based guidance at the point of care, these tools could help reduce diagnostic errors and improve patient outcomes. Local deployment ensures data protection for patient health information while eliminating reliance on external searches, addressing both privacy concerns and workflow efficiency.
Specific clinical applications demonstrate practical utility: (1) real-time emergency interpretation for distinguishing ST-elevation myocardial infarction (STEMI) mimics from true infarction, (2) continuous monitoring integration for intelligent alarm management and arrhythmic event prediction \cite{Khera2024AICardiology}, and (3) rapid processing of large-scale ECG databases for novel biomarker identification \cite{losch2025finetuningllmssmallmedical}. Supporting evidence shows promise across cardiology tasks, with ChatGPT-4o achieving 75.9\% accuracy on cardiology board questions \cite{Malik2025ChatGPT4oCardiology}.
These models could also integrate into advanced systems, including multimodal diagnostic frameworks or agentic systems for context-aware clinical reasoning, such as VLMs for ECG graph interpretation. Systems like ECG-Chat demonstrate this capability, combining ECG signal processing with textual analysis \cite{zhao2025ecgchatlargeecglanguagemodel}.
However, clinical deployment requires extensive validation including real-world testing, safety evaluations, workflow integration, and regulatory approval. At this stage, these models are better suited for augmenting clinical practice by helping clinicians find information rather than making autonomous decisions.

%%%%%%%%%%%%%%%%%%%%%%%%%%%%%%%%%%%%%%%%%%%%%%%%%%%
%%%%%%%%%%%%%%%%%%%%%%%%%%%%%%%%%%%%%%%%%%%%%%%%%%%
%%%%%%%%%%%%%%%%%%%%%%%%%%%%%%%%%%%%%%%%%%%%%%%%%%%
\section{Methods}\label{sec:methods}

\subsection{Finetuning (FT)}

We generated question-answer pairs by prompting an LLM with literature context, similar to \cite{zheng2025miriadaugmentingllmsmillions} (full prompt in Appendix~\ref{ap:promptData}). Starting from PDF files, we converted them to markdown using MinerU \cite{wang2024mineruopensourcesolutionprecise}, then cleaned the files by removing author information, tables of contents, references, and acknowledgements as in \cite{SchillingWilhelmi.24.07.2024}. We split files by chapters (maximum 10 chapters and 50,000 tokens estimated with TikToken \cite{openai_tiktoken}) and prompted Llama 3.3 70B to generate Q\&A pairs. We selected Llama 3.3 70B as an open-source model that consistently generated context-aligned questions without incorporating external knowledge, verified using AlignScore \cite{zha-etal-2023-alignscore}.

%\heading{Considered Base Models}
We finetuned Llama 3.1 70B and 8B models and compared them against their base versions and Claude Sonnet 3.7, representing state-of-the-art performance.
%\heading{finetuning Methodology}
We used the Hugging Face Trainer with QLoRA for memory-efficient training \cite{wolf-etal-2020-transformer, dettmers2023qloraefficientfinetuningquantized}. The dataset was split 80/10/10 (train/validation/test) per file, with essential knowledge labeled by the human expert fully integrated into training. Prompts followed Llama 3 specifications with role tokens, bos, and eos tokens \cite{dubey2024llama3herdmodels}, with loss computed only on answers. We used AdamW optimizer (paged-32) \cite{marie2024fine}, cosine learning-rate scheduler \cite{raschka2024lora}, batch size of eight, and cross-entropy loss \cite{wolf-etal-2020-transformer}. LoRA parameters were applied to attention, feedforward, and output layers \cite{Hu.17.06.2021}, resulting in 3.7\% trainable parameters. Testing multiple configurations yielded optimal performance at $r=256,\alpha=128$ \cite{raschka2024lora}. We monitored performance using domain-specific multiple-choice metrics (10\% per file). Training loss decreased initially but validation loss increased after two epochs while multiple-choice accuracy stabilized, prompting us to stop training at two epochs. Higher $\alpha$ values (512) reduced 8B model accuracy. Experiments with Llama 3.1 70B on H100 GPUs showed similar patterns.

\subsection{Retrieval-augmented generation (RAG)}
%\heading{General Approach}
RAG offers another approach for domain question-answering \cite{ong2024surgeryllm} and reduces LLM hallucination when knowledge is missing \cite{huang2023survey}. The RAG process involves several steps: First, documents converted to markdown using MinerU are split and embedded using a selected splitting strategy, vector database, and embedding model. During retrieval, the user's message performs a similarity search on the vector database, selecting the top-k matching chunks as context. In augmentation, a prompt combines the user message and context, which is then passed to the LLM for response generation \cite{Kamath.2024, lewis2020retrieval}.

%\heading{RAG Setup}
We tested multiple configurations (detailed in Appendix~\ref{ap:RAGconfig}). The optimal configuration uses recursive splitting with 1024-token chunks and 100-token overlap \cite{theja2023chunk}, PubMedBERT embeddings \cite{neuml_pubmedbert_embeddings}, top-20 retrieval, and reranking with top-5, achieving first rank across all subsets. Reranking embeds document parts and queries simultaneously, enabling more precise similarity ranking \cite{ampazis2024improving}. We implemented this RAG configuration for Llama 3.1 8B Instruct and Llama 3.1 70B Instruct models.

\subsection{Evaluation methodology} 
\begin{figure}[htbp] % htbp = here, top, bottom, page
  \centering
  \includegraphics[]{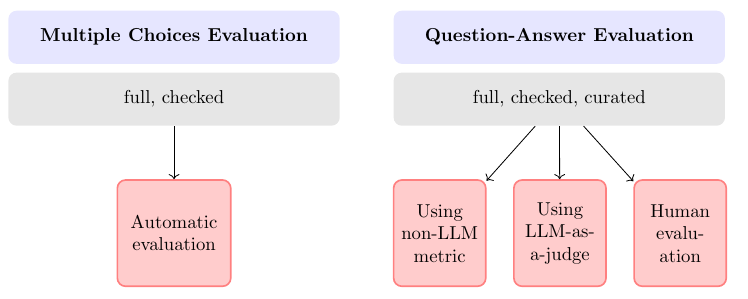}
  \caption{Evaluation layers in the evaluation approach} 
  \label{fig:eval} 
\end{figure}

We assess the quality of the generated content by means of four qualitatively different evaluation methods, see Figure \ref{fig:eval} for an overview, which we describe below.

\paragraph{Multiple-choice evaluation}
Evaluating LLM knowledge requires a multi-faceted approach \cite{Teo2025GenerativeAI}. While multiple-choice tests can assess LLM understanding \cite{hendrycks2021measuringmassivemultitasklanguage, Haghighi.}, and general medical tests exist on Hugging Face \cite{ura2024openmedicalllm}, these are insufficient for evaluating specialized ECG knowledge. Therefore, we generated a domain-specific multiple-choice test from the provided medical documents using the data generation process for generating the finetuning data, adjusting the prompt to generate multiple-choice questions with answer options instead of question-answer pairs.

\paragraph{Text similarity metrics}
Multiple-choice tests cannot represent real-world scenarios where LLMs answer open-ended questions. Consequently, a comprehensive evaluation should include both factual knowledge assessments and open-ended question evaluations. An evaluation based on automatic evaluation metrics supplements open-ended response evaluation despite uncertain quality assessment. Metrics like BLEU \cite{bleu_score} and ROUGE \cite{lin2004rouge}, designed for summarization and translation, are suboptimal for question-answering but can assess keyword presence in responses. BERTScore \cite{bertscore} evaluates semantic meaning using contextual embeddings, accommodating synonymous expressions, though its quality depends on text formulation and cannot account for semantically valid alternative interpretations beyond reference answers. In our evaluation approach, we employ all of these metrics to provide a comprehensive evaluation of model responses. 

\paragraph{LLM-as-a-judge}
For open-ended questions, the answers are also assessed using LLM-as-a-judge \cite{Kim.02.05.2024}. We selected Deepseek R1 as the evaluator due to its strong reasoning capabilities \cite{deepseekai2025deepseekr1incentivizingreasoningcapability}, making it effective for comparing responses and assessing accuracy. Deepseek R1 functions only as an evaluator and is not itself evaluated. The judge LLM focuses solely on correctness rather than ranking or scoring. Most importantly, we include the context that was used to originally generate the question in order to mitigate hallucinations \cite{huang2023survey}

\paragraph{Human evaluation}
Human evaluation is carried out as a very valuable evaluation to confirm the other assessments. For this purpose, we collected the answers of finetuned models and state-of-the-art models to the questions provided by the medical expert and handed them over to the medical expert. The expert assesses whether the respective answers are correct, correct but incomplete, incorrect, or partially incorrect. We decided not to assign points to the answers themselves, as the allocation of points could be too variable and the criteria for evaluation could vary unconsciously. This process is carried out for two distinct subsets of questions provided by the human expert. Human evaluation is resources are limited and it is time-consuming \cite{li2024crowdsourceddatahighqualitybenchmarks}.

\paragraph{Statistical significance testing}
Statistical fluctuations may obscure true method rankings. We use empirical bootstrapping ($n=1000$ iterations) to assess statistical significance. For each evaluation mode, we used respective scores as input; for human evaluation, we converted categorical answers to numerical scores (correct: 1; partially correct: 0.75; partially incorrect: 0.25; incorrect: 0). We performed pairwise comparisons by bootstrapping score differences between models. Performance differences are considered statistically significant if the 95\% confidence interval excludes zero. This procedure generates ranked lists with ties, where ties indicate no statistically significant performance difference. This framework provides statistically robust, unified comparisons across all evaluation modes.

\paragraph{Evaluation datasets}
For question-answer evaluation, we used a subset from the data generation process, split into train-validation-test sets (80/10/10), with test data serving as the evaluation dataset. While LLM-generated data lacks verified quality, it provides high quantity. These datasets (question-answer pairs and multiple-choice questions) are labeled as \textit{full}. A medical expert corrected subsets of 534 multiple-choice and 537 question-answer samples, labeled as \textit{checked}. For human evaluation, the expert additionally provided 47 questions, labeled as \textit{curated}.

\backmatter

\section*{Declarations}

\subsection*{Conflict of interest}
The authors declare no competing interests.
\subsection*{Data availability}
We release the human evaluation dataset along with model predictions and expert evaluations. The dataset used for pretraining and automatic evaluation cannot be released to align with §60d UrhG (implementing EU Directive 2019/790). The same holds for the model weights of the finetuned model.
\subsection*{Code availability}
The complete source code for preprocessing, finetuning and retrieval augmented generation is available at \url{https://github.com/AI4HealthUOL/ecg-llm}.
\subsection*{Author contributions}
Conceptualization: NS, WH; Methodology: NS, LA; Software: LA; Validation: LA, NS; Data Curation: WH; Visualization: LA; Writing - Original Draft: LA, NS; Writing - Review: LA, NS, WH; Supervision: NS

%%===========================================================================================%%
%% If you are submitting to one of the Nature Portfolio journals, using the eJP submission   %%
%% system, please include the references within the manuscript file itself. You may do this  %%
%% by copying the reference list from your .bbl file, paste it into the main manuscript .tex %%
%% file, and delete the associated \verb+\bibliography+ commands.                            %%
%%===========================================================================================%%
\bibliography{bibfile}% common bib file
%% if required, the content of .bbl file can be included here once bbl is generated
%%\input sn-article.bbl

\clearpage
\begin{appendices}
%\section{Appendix} Otherwise it is Appenix A Appendix
\lstdefinestyle{python}{
    language=Python,
    basicstyle=\ttfamily\small,
    keywordstyle=\color{blue},
    stringstyle=\color{red},
    commentstyle=\color{green!70!black},
    numbers=left,
    numberstyle=\tiny\color{gray},
    stepnumber=1,
    numbersep=10pt,
    backgroundcolor=\color{gray!10},
    showspaces=false,
    showstringspaces=false,
    showtabs=false,
    frame=single,
    rulecolor=\color{black},
    tabsize=4,
    captionpos=b,
    breaklines=true,
    breakatwhitespace=true,
    escapeinside={(*@}{@*)},
    morekeywords={self},
}
\section{Prompt data generation}\label{ap:promptData}
Listing \ref{list:final_prompt} shows the prompt for the generation of Q\&A-pairs
\lstset{literate={’}{{\textquotesingle}}1}
\lstset{literate={’}{{\textquotesingle}}1}  \lstset{literate={’}{{\textquotesingle}}1} \begin{lstlisting}[caption=Prompt to generate question and answer pairs, label=list:final_prompt, style=python]
system_message = f"""You are a Teacher/ Professor in the medical field. Your task is to setup a examination with free text answers. Using the provided context, formulate questions with different difficulties that capture the medical content from the context. Please also give the answers, so that the test could be corrected afterwards.
If you cannot generate any question to medical content, please skip it.
You MUST obey the following criteria:
- Restrict the question to the context information provided.
- vary between different question words (e.g. what, why, which, how, etc.)
- Ensure every question is fully self-contained and answerable without requiring additional context or prior questions/answers.
- Do NOT ask for figures, algortihms, tables, names of the present study or similar.
- Do NOT put phrases like "given provided context" or "in this work" or "in this case" or "what is algorithm a" or some questions regarding the study
- Replace these terms with specific details
- ONLY ask for medical details. DO NOT ask about the study, author or index, IGNORE them

BAD questions:
- What are the symptoms of the disease described
- How many patients were included in the study

GOOD questions:
- What are the symptoms of Corona
- Why should the patient drink so much water when having a fever

Sometimes the context may contain overhead such as titles, authors, study information or similar. Please only use the content that has a medical context. 

Output: ONLY JSON."""

user_message = f"""Here is the context for generating medical questions:

{context}
{self.parser.get_format_instructions()}
Please answer in English and do not use any German words or phrases. Translate technical terms into English if necessary.
"""
\end{lstlisting}

\section{Prompt generating multiple choices}\label{ap:promptMultipleCHoice}
Listing 
\ref{list:final_prompt_multiple} shows the prompt for generating multiple choices.
\lstset{literate={’}{{\textquotesingle}}1}
\lstset{literate={’}{{\textquotesingle}}1}  \lstset{literate={’}{{\textquotesingle}}1} \begin{lstlisting}[caption=Prompt to generate multiple-choices, label=list:final_prompt_multiple, style=python]
system_message = f"""You are a Teacher/Professor in the medical field. 
Your task is to create multiple-choice questions with different difficulties based on the medical context provided.

For each question:
- Generate one question about important medical content
- Create exactly 4 answer options
- One option must be correct and from the context
- Three options must be plausible but incorrect (make these up)
- All options should have similar length and style
- Focus only on medical content
If you cannot generate any question to medical content, please skip it.
You MUST obey the following criteria:
- Restrict the question to the context information provided.
- vary between different question words (e.g. what, why, which, how, etc.)
- Ensure every question is fully self-contained and answerable without requiring additional context or prior questions/answers.
- Do NOT ask for figures, algortihms, tables, names of the present study or similar.
- Do NOT put phrases like "given provided context" or "in this work" or "in this case" or "what is algorithm a" or some questions regarding the study
- Replace these terms with specific details
- ONLY ask for medical details. DO NOT ask about the study, author or index, IGNORE them

BAD questions:
- What are the symptoms of the disease described
- How many patients were included in the study

GOOD questions:
- What are the symptoms of Corona
- Why should the patient drink so much water when having a fever

Sometimes the context may contain overhead such as titles, authors, study information or similar. Please only use the content that has a medical context. 

Output: ONLY JSON."""

user_message = f"""Generate multiple-choice questions from this context:
{context}
{parser.get_format_instructions()}
Please answer in English and do not use any German words or phrases. Translate technical terms into English if necessary.

"""
\end{lstlisting}
\section{Results of different RAG-configurations}\label{ap:RAGconfig}
We considered various RAG configurations before selecting the final one. Therefore, we evaluated the multiple-choice results on the different subsets with different configurations. For splitting the documents, we consider the Markdown splitting approach as implemented in Langchain as the documents are converted as markdown files, and a recursive chunking strategy with a size of 1024 with 100 overlap suggested by \cite{theja2023chunk}. For the embedding process, we selected Chroma as the vector database and compared two embedding models: multilingual-e5-large-instruct as one of the leading embedding models in the Hugging Face leaderboard for embedding models \cite{huggingfaceMTEB2025}, and PubMedBERT \cite{pubmedbert} for its specialization in biomedical text.
Table \ref{tab:rag-combined-ranked} shows the results of the  different RAG configurations across splitting strategies, retrieval settings, and embedding models, focusing on factual accuracy in multiple-choice evaluations. Overall, recursive splitting with 1,024-token chunks, larger top-k values, and reranking achieves the best performance, while both embedding models performed similarly, with PubMedBERT showing slightly better performance in the checked subset.
\begin{table}[htbp]
\centering
\begin{tabular}{lllccc}
\toprule
\textbf{Splitting} & \textbf{Top-k} & \textbf{Embedding} &  
\textbf{Full} & \textbf{Checked} & \textbf{English} \\
\midrule
\makecell{\textbf{Recursive}\\\textbf{(1,024 + 100)}} & \makecell{\textbf{20 + Reranking}\\\textbf{top 5}} & \textbf{PubMedBERT} & \textbf{88.5\% (1)} & \textbf{85.0\% (1)} & \textbf{88.3\% (1)} \\
\makecell{Recursive\\(1,024 + 100)} & \makecell{20 + Reranking\\top 5} & Multilingual & 88.8\% (1) & 83.9\% (2) & 88.5\% (1) \\
\makecell{Recursive\\(1,024 + 100)} & \makecell{10 + Reranking\\top 5} & PubMedBERT & 87.3\% (3) & 83.9\% (3) & 86.9\% (5) \\
\makecell{Recursive\\(1,024 + 100)} & \makecell{10 + Reranking\\top 5} & Multilingual & 87.6\% (3) & 82.8\% (4) & 87.2\% (3) \\
\makecell{Recursive\\(1,024 + 100)} & 10 & Multilingual & 87.5\% (3) & 82.7\% (4) & 87.2\% (3) \\
\makecell{Recursive\\(1,024 + 100)} & 5 & PubMedBERT & 86.7\% (6) & 82.0\% (6) & 86.3\% (6) \\
\makecell{Recursive\\(1,024 + 100)} & 5 & Multilingual & 86.7\% (7) & 81.1\% (8) & 86.5\% (7) \\
\makecell{Recursive\\(1,024 + 100)} & 10 & PubMedBERT & 86.5\% (8) & 82.2\% (6) & 86.1\% (8) \\
\makecell{Recursive\\(1,024 + 100)} & 3 & Multilingual & 85.9\% (9) & 80.5\% (9) & 85.7\% (9) \\
Markdownheader & 2 & PubMedBERT & 85.2\% (10) & 79.4\% (10) & 84.7\% (10) \\
Markdownheader & 2 & Multilingual & 84.4\% (11) & 76.2\% (12) & 84.0\% (11) \\
\makecell{Recursive\\(1,024 + 100)} & 3 & PubMedBERT & 83.9\% (12) & 78.7\% (11) & 83.7\% (11) \\\bottomrule
\end{tabular}
\caption{Combined and ranked comparison of RAG configurations showing multiple-choice accuracy with corresponding rank across subsets.}
\label{tab:rag-combined-ranked}
\end{table}

\section{LLM-as-a-judge evaluation}
\label{app:llmaaj}
To verify the LLM-as-a-judge evaluation, we gave a smaller, random subset of correct and incorrect labeled answers from one of the finetuned models (LLama 3.1 8B + FT, $r=64, \alpha=16$) to the human expert. An overview of the results is shown in Table~\ref{tab:confirmLLMJudge}. The correct answers are also labeled as correct. From the wrong answers, three answers are labeled as correct, one as partially correct.

\begin{table}[h]
    \centering
    \caption{Validation of LLM-as-a-judge evaluation through human expert review. Values represent counts of responses. Agreement between automated and human evaluation was 80\% (16/20 cases). Cases labeled as incorrect by the LLM judge showed heterogeneous human assessments, highlighting the complexity of response quality evaluation.}
    \label{tab:confirmLLMJudge}
    \begin{tabular}{@{}lrrrr@{}}
        \toprule
        \textbf{LLM-as-a-Judge} & \textbf{Total} & \textbf{Human:} & \textbf{Human:} & \textbf{Human:}  \\
        \textbf{Label} & \textbf{Count} & \textbf{Incorrect} & \textbf{Correct} & \textbf{Partially Correct} \\
        \midrule
        Correct & 13 & 0 & 13 & 0  \\
        Incorrect & 7 & 3 & 3 & 1  \\
        \bottomrule
    \end{tabular}
\end{table}
\clearpage
\section{Results of other tested configurations}\label{ap:otherTestedConfig}
\begin{figure}[h]
    \centering
    \includegraphics[]{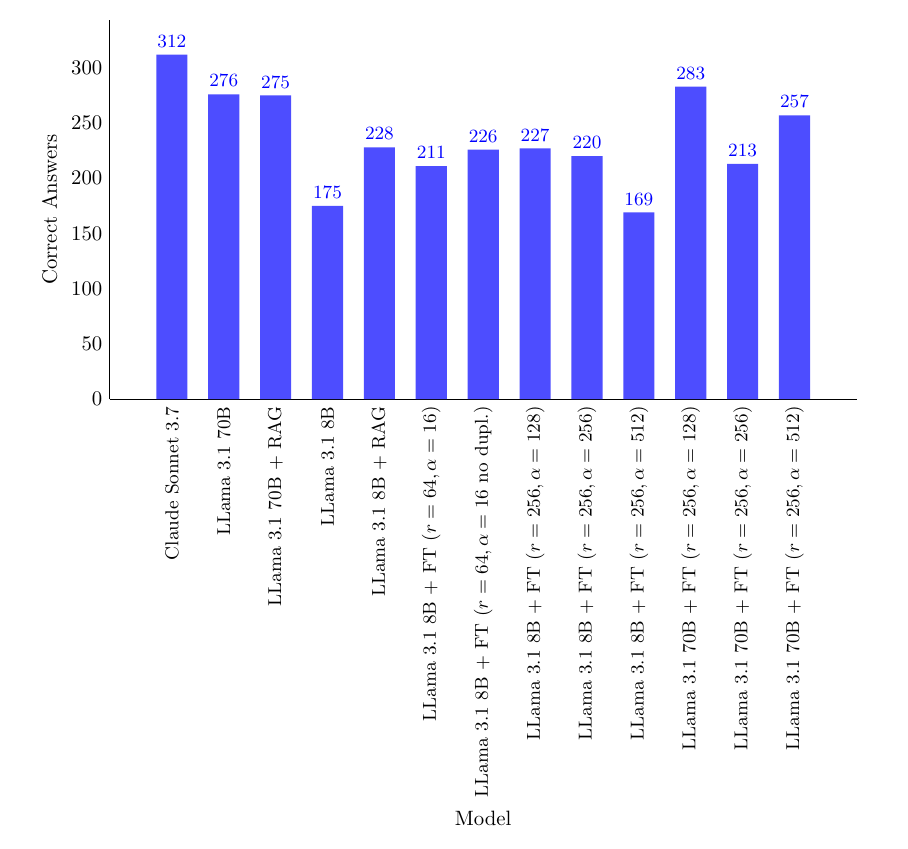}
    \caption{Correct answers resulting from 416 questions evaluated with LLM-as-a-judge}
    \label{fig:llmjudgeall}
\end{figure}

\begin{figure}[h]
    \centering
    \includegraphics[]{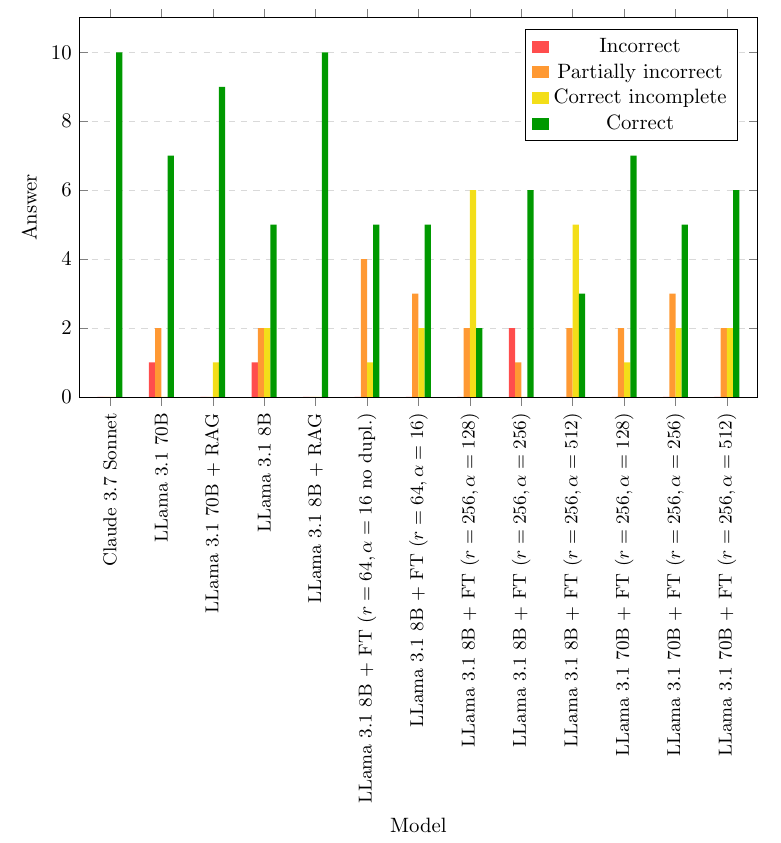}
    \caption{Human evaluation on the ten factual questions}
    \label{fig:human10_all}
\end{figure}

\begin{figure}[h]
    \centering
    \includegraphics[]{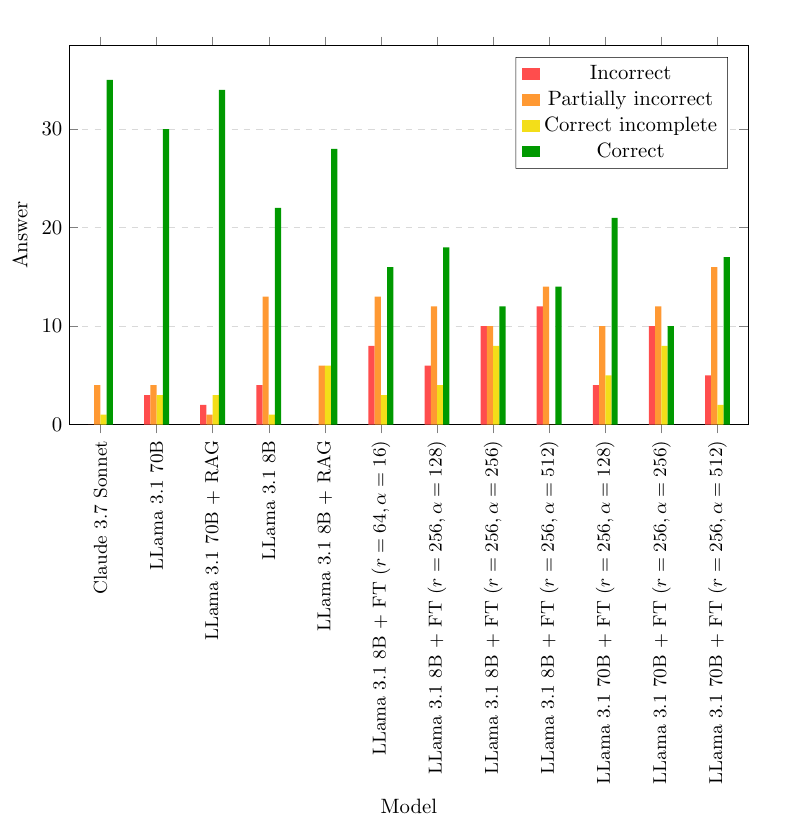}
    \caption{Human evaluation on the 40 semantically complex questions}
    \label{fig:human40_all}
\end{figure}

We additionally finetuned models using various LoRA configurations. Specifically, we tested configurations with $\alpha \in {128, 256, 512}$ and a the corresponding rank of $r = 256$. Furthermore, we evaluated a setup with $r = 64$ and $\alpha = 16$, as suggested by \cite{li-etal-2024-llamacare}. One experiment was also conducted using a training dataset from which all duplicated questions were removed. The results of these experiments are presented in the following Tables \ref{tab:acc_results_full_all}, \ref{tab:bleu_rouge1_all}, \ref{tab:rouge2_rougel}, \ref{tab:bertscore} and Figures \ref{fig:llmjudgeall}, \ref{fig:human10_all}, \ref{fig:human40_all}. For the 8B model, the configuration with $\alpha = 512$ failed to run successfully. Overall, the configuration with $r = 256$ and $\alpha = 128$ achieved the best performance. As initial experiments without duplicated questions resulted in worse performance compared to those including some duplicates, further testing under that condition was discontinued. The overall ranking is shown in Section \ref{ap:rankall}

\paragraph{Multiple choice evaluation}
Table \ref{tab:acc_results_full} presents the multiple-choice results for all finetuned configurations.

\begin{table}[h]
\centering
\begin{tabular}{p{5.5cm}cccc}
\toprule
\textbf{Model} &
\makecell{\textbf{Special}} &
\makecell{\textbf{Full}} &
\makecell{\textbf{Checked}}\\
\midrule
{LLama 3.1 8B Instruct} & 73.3\% & 81.6\% & 73\%  \\
{LLama 3.1 70B Instruct} & 82.4\% &{ 88.2\%} & 81.5\%  \\
{Claude 3.7 Sonnet Latest} & 82.3\% & 88\% & 81.7\%  \\
{LLama 3.1 8B + FT (r= 64, $\alpha=16$)} & 83.7\% & 87.9\% & 83\%  \\
{LLama 3.1 8B + FT (r= 64, $\alpha=16$ no duplicates)} & 83.9\% & 87.6\% & 81.5\%  \\
{LLama 3.1 8B + FT ($r=256, \alpha=128$)} & 84.9\% & 87.5\% & {84.3\%} \\
{LLama 3.1 8B + FT ($r=256, \alpha=256$)} & 84.2\% & 85\% & 84.1\% \\
{LLama 3.1 8B + FT ($r=256, \alpha=512$)} & 79.3\% & 79.2\% & 80\%  \\
{LLama 3.1 70B + FT ($r=256, \alpha=128$)} & 90.2\% & 92\% & 88.2\%\\
{LLama 3.1 70B + FT ($r=256, \alpha=256$)} & 87.4\% & 88.4\% & 84.8\%\\
{LLama 3.1 70B + FT ($r=256, \alpha=512$)} & 89.2\% & 92.1\% & 89.7\% \\
{LLama 3.1 8B  + {RAG}} & 86.1\% & 88.5\% & 85\% \\
{Llama 3.1 70B + RAG} & 87.9\% & 90.2\%  & 86.5\%  \\
\bottomrule
\end{tabular}
\caption{Performance of different models across different multiple-choice sets}
\label{tab:acc_results_full_all}
\end{table}

\paragraph{BLEU, ROUGE, BERTScore evaluation}
Table \ref{tab:bleu_rouge1_all} presents the BLEU and ROUGE-1 values, table \ref{tab:rouge2_rougel} the ROUGE-2 and ROUGE-L and Table \ref{tab:bertscore} the BERTScores for all finetuned configurations.
\begin{table}[h]
\centering
\begin{tabular}{p{5.5cm}cccc}
\toprule
\textbf{Model} & \textbf{BLEU} & \textbf{R1 F1} & \textbf{R1 Precision} & \textbf{R1 Recall} \\
\midrule
{LLama 3.1 8B + FT ($r=64, \alpha=16$)} & 0.1103 & 0.4031 & 0.4035 & 0.4549 \\
{LLama 3.1 8B + FT ($r=64, \alpha=16$ no duplicates)} & 0.1083 & 0.4002 & 0.3984 & 0.4545 \\
{LLama 3.1 8B + FT ($r=256, \alpha=128$)} & 0.1195 & 0.4063 & 0.4039 & 0.4597 \\
{LLama 3.1 8B + FT ($r=256, \alpha=256$)} & 0.1196 & 0.4030 & 0.3996 & 0.4553 \\
{LLama 3.1 8B + FT ($r=256, \alpha=512$)} & 0.1149 & 0.3910 & 0.3898 & 0.4396 \\
{LLama 3.1 70B + FT ($r=256, \alpha=128$)} & 0.1289 & 0.4270 & 0.4292 & 0.4749 \\
{LLama 3.1 70B + FT ($r=256, \alpha=256$)} & 0.1355 & 0.4162 & 0.4168 & 0.4617 \\
{LLama 3.1 70B + FT ($r=256, \alpha=512$)} & 0.1281 & 0.4266 & 0.4262 & 0.4781 \\
{LLama 3.1 8B Instruct} & 0.0788 & 0.3537 & 0.2903 & 0.5252 \\
{Claude 3.7 Sonnet Latest} & 0.0694 & 0.3661 & 0.3247 & 0.4897 \\
{LLama 3.1 70B Instruct} & 0.0962 & 0.3852 & 0.3307 & 0.5347 \\
{LLama 3.1 8B Instruct + RAG} & 0.0763 & 0.3481 & 0.2856 & 0.5162 \\
{LLama 3.1 70B Instruct + RAG} & 0.0339 & 0.2414 & 0.1668 & 0.5488 \\
\bottomrule
\end{tabular}
\caption{{BLEU} and {ROUGE}-1 scores for different models on test dataset}
\label{tab:bleu_rouge1_all}
\end{table}

\begin{table}[h]
\centering
\begin{tabular}{p{5cm}cccccc}
\toprule
\textbf{Model} & \textbf{R2 F1} & \textbf{R2 P} & \textbf{R2 R} & \textbf{RL F1} & \textbf{RL P} & \textbf{RL R} \\
\midrule
{LLama 3.1 8B + FT ($r=64, \alpha=16$)} & 0.2197 & 0.2202 & 0.2483 & 0.3518 & 0.3523 & 0.3979 \\
{LLama 3.1 8B + FT ($r=64, \alpha=16$) no duplicates} & 0.2164 & 0.2149 & 0.2465 & 0.3482 & 0.3466 & 0.3967 \\
{LLama 3.1 8B + FT ($r=256, \alpha=128$)} & 0.2279 & 0.2263 & 0.2582 & 0.3568 & 0.3546 & 0.4046 \\
{LLama 3.1 8B + FT ($r=256, \alpha=256$)} & 0.2262 & 0.2239 & 0.2559 & 0.3538 & 0.3506 & 0.4004 \\
{LLama 3.1 8B + FT ($r=256, \alpha=512$)} & 0.2167 & 0.2153 & 0.2436 & 0.3443 & 0.3429 & 0.3878 \\
{LLama 3.1 70B + FT ($r=256, \alpha=128$)} & 0.2449 & 0.2455 & 0.2725 & 0.3764 & 0.3782 & 0.4192 \\
{LLama 3.1 70B + FT ($r=256, \alpha=256$)} & 0.2426 & 0.2424 & 0.2696 & 0.3694 & 0.3699 & 0.4103 \\
{LLama 3.1 70B + FT ($r=256, \alpha=512$)} & 0.2445 & 0.2437 & 0.2739 & 0.3761 & 0.3755 & 0.4226 \\
{LLama 3.1 8B Instruct} & 0.1730 & 0.1429 & 0.2586 & 0.2906 & 0.2382 & 0.4367 \\
{Claude 3.7 Sonnet Latest} & 0.1685 & 0.1502 & 0.2297 & 0.2992 & 0.2650 & 0.4054 \\
{LLama 3.1 70B Instruct} & 0.1981 & 0.1715 & 0.2766 & 0.3208 & 0.2750 & 0.4503 \\
{LLama 3.1 8B Instruct + RAG} & 0.1689 & 0.1397 & 0.2514 & 0.2831 & 0.2320 & 0.4250 \\
{LLama 3.1 70B Instruct + RAG} & 0.1064 & 0.0734 & 0.2512 & 0.1871 & 0.1288 & 0.4382 \\\bottomrule

\end{tabular}
\caption{{ROUGE}-2 and {ROUGE}-L scores for different models on test dataset}
\label{tab:rouge2_rougel}
\end{table}

\begin{table}[h]
\centering
\begin{tabular}{p{5.5cm}ccc}
\toprule
\textbf{Model} & \textbf{Precision} & \textbf{Recall} & \textbf{F1} \\
\midrule
{LLama 3.1 8B + FT ($r=64, \alpha=16$)} & 0.3644 & 0.3693 & 0.3665 \\
{LLama 3.1 8B + FT ($r=64, \alpha=16$) no duplicates} & 0.3610 & 0.3670 & 0.3636 \\
{LLama 3.1 8B + FT ($r=256, \alpha=128$)} & 0.3660 & 0.3748 & 0.3700 \\
{LLama 3.1 8B + FT ($r=256, \alpha=256$)} & 0.3599 & 0.3710 & 0.3651 \\
{LLama 3.1 8B + FT ($r=256, \alpha=512$)} & 0.3537 & 0.3573 & 0.3552 \\
{LLama 3.1 70B + FT ($r=256, \alpha=128$)} & 0.3884 & 0.3932 & 0.3904 \\
{LLama 3.1 70B + FT ($r=256, \alpha=256$)} & 0.3732 & 0.3826 & 0.3776 \\
{LLama 3.1 70B + FT ($r=256, \alpha=512$)} & 0.3846 & 0.3961 & 0.3900 \\
{LLama 3.1 8B Instruct} & 0.2194 & 0.3699 & 0.2936 \\
{Claude 3.7 Sonnet Latest} & 0.2703 & 0.4112 & 0.3395 \\
{LLama 3.1 70B Instruct} & 0.2795 & 0.4180 & 0.3476 \\
{LLama 3.1 8B Instruct + RAG} & 0.2157 & 0.3584 & 0.2861 \\
{LLama 3.1 70B Instruct + RAG} & 0.2811 & 0.1418 & 0.0095 \\
\bottomrule
\end{tabular}
\caption{BERTScore of the models over test dataset}
\label{tab:bertscore}
\end{table}

\paragraph{LLM-as-a-judge evaluation}
Figure \ref{fig:llmjudgeall} shows the results of LLM-as-a-judge for all finetuned configurations.

\paragraph{Human evaluation}
Figures \ref{fig:human10_all} shows the results of all finetuned models on the 10 questions asking for facts. Figure \ref{fig:human40_all} shwos the results of all finetuned models on the 40 more complex questions, evaluated by an human expert.

\section{Overall ranking}\label{ap:rankall}
Table \ref{tab:ranks_all} shows all models with their corresponding ranks in the different evaluations.
\colorlet{best}{green!80!black}
\colorlet{tempcolor}{green!50!yellow}

\colorlet{good}{tempcolor!90!brown}
\colorlet{medium}{yellow!80!brown}
\colorlet{bad}{orange!80}
\colorlet{worst}{red!70}

\newcommand{\rankcolor}[1]{%
    \ifnum#1=1 \cellcolor{best}{#1}%
    \else\ifnum#1=2 \cellcolor{good}{#1}%
    \else\ifnum#1=3 \cellcolor{medium}{#1}%
    \else\ifnum#1=4 \cellcolor{medium}{#1}%
    \else\ifnum#1<8 \cellcolor{bad}{#1}%
    \else \cellcolor{worst}{#1}%
    \fi\fi\fi\fi\fi
}

\begin{table}[htbp]
\centering
\rotatebox{90}{%
\begin{tabular}{|p{2.7cm}|p{1cm}|p{1.4cm}|p{1.4cm}|c|c|c|c|p{1.2cm}|p{1cm}|p{1.7cm}|p{1.7cm}|}
\hline
\textbf{Model} &
\textbf{Full mult. choi.} &
\textbf{Special multiple choices} &
\textbf{Checked multiple choices} &
\textbf{{BLEU}} &
\textbf{R-1} &
\textbf{R-2} &
\textbf{R-L} &
\textbf{BERT\-Score} &
\textbf{LLM-as-a-judge} &
\textbf{Human 10 questions} &
\textbf{Human 40 questions}\\
\hline
LLama 3.1 8B & \rankcolor{12} & \rankcolor{12} & \rankcolor{12} & \rankcolor{10} & \rankcolor{11} & \rankcolor{10} & \rankcolor{11} & \rankcolor{11} & \rankcolor{11} & \rankcolor{7} & \rankcolor{5} \\
\hline
LLama 3.1 70B & \rankcolor{6} & \rankcolor{7} & \rankcolor{9} & \rankcolor{9} & \rankcolor{9} & \rankcolor{9} & \rankcolor{9} & \rankcolor{9} & \rankcolor{3} & \rankcolor{5} & \rankcolor{1} \\
\hline
Claude 3.7 Sonnet & \rankcolor{6} & \rankcolor{7} & \rankcolor{9} & \rankcolor{12} & \rankcolor{10} & \rankcolor{11} & \rankcolor{10} & \rankcolor{10} & \rankcolor{1} & \rankcolor{1} & \rankcolor{1} \\
\hline
LLama 3.1 8B + FT (r=64, $\alpha=16$)& \rankcolor{6} & \rankcolor{7} & \rankcolor{5} & \rankcolor{7} & \rankcolor{6} & \rankcolor{6} & \rankcolor{5} & \rankcolor{5} & \rankcolor{5} & \rankcolor{7} & \rankcolor{8} \\
\hline
LLama 3.1 8B + FT (r=64, $\alpha=16$ no duplicates) & \rankcolor{9} & \rankcolor{7} & \rankcolor{9} & \rankcolor{7} & \rankcolor{5} & \rankcolor{7} & \rankcolor{7} & \rankcolor{6} & \rankcolor{6} & \rankcolor{7} &  \\
\hline
LLama 3.1 8B + FT ($r=256, \alpha=128$)& \rankcolor{9} & \rankcolor{5} & \rankcolor{5} & \rankcolor{4} & \rankcolor{4} & \rankcolor{4} & \rankcolor{4} & \rankcolor{4} & \rankcolor{5} & \rankcolor{7} & \rankcolor{5} \\
\hline
LLama 3.1 8B + FT ($r=256, \alpha=256$) & \rankcolor{11} & \rankcolor{7} & \rankcolor{5} & \rankcolor{4} & \rankcolor{6} & \rankcolor{4} & \rankcolor{6} & \rankcolor{6} & \rankcolor{5} & \rankcolor{13} & \rankcolor{10} \\
\hline
LLama 3.1 8B + FT ($r=256, \alpha=512$) & \rankcolor{12} & \rankcolor{12} & \rankcolor{12} & \rankcolor{6} & \rankcolor{8} & \rankcolor{8} & \rankcolor{7} & \rankcolor{8} & \rankcolor{11} & \rankcolor{12} & \rankcolor{10} \\
\hline
LLama 3.1 70B + FT ($r=256, \alpha=128$) & \rankcolor{1} & \rankcolor{1} & \rankcolor{1} & \rankcolor{1} & \rankcolor{1} & \rankcolor{1} & \rankcolor{1} & \rankcolor{1} & \rankcolor{2} & \rankcolor{4} & \rankcolor{5} \\
\hline
LLama 3.1 70B + FT ($r=256, \alpha=256$) & \rankcolor{5} & \rankcolor{4} & \rankcolor{3} & \rankcolor{1} & \rankcolor{3} & \rankcolor{1} & \rankcolor{3} & \rankcolor{3} & \rankcolor{5} & \rankcolor{7} & \rankcolor{10} \\
\hline
LLama 3.1 70B + FT ($r=256, \alpha=512$) & \rankcolor{2} & \rankcolor{1} & \rankcolor{1} & \rankcolor{3} & \rankcolor{1} & \rankcolor{1} & \rankcolor{1} & \rankcolor{1} & \rankcolor{3} & \rankcolor{5} & \rankcolor{8} \\
\hline
LLama 3.1 8B + {RAG} & \rankcolor{4} & \rankcolor{5} & \rankcolor{5} & \rankcolor{11} & \rankcolor{12} & \rankcolor{11} & \rankcolor{12} & \rankcolor{12} & \rankcolor{5} & \rankcolor{1} & \rankcolor{1} \\
\hline
LLama 3.1 70B + {RAG} &  \rankcolor{2}& \rankcolor{3} & \rankcolor{3} & \rankcolor{13} & \rankcolor{13} & \rankcolor{13} & \rankcolor{13} & \rankcolor{13} & \rankcolor{3} & \rankcolor{1} & \rankcolor{1} \\
\hline
\end{tabular}
}
\caption{Ranks of different models across the different evaluation methods}
\label{tab:ranks_all}
\end{table}

% An appendix contains supplementary information that is not an essential part of the text itself but which may be helpful in providing a more comprehensive understanding of the research problem or it is information that is too cumbersome to be included in the body of the paper.

%%=============================================%%
%% For submissions to Nature Portfolio Journals %%
%% please use the heading ``Extended Data''.   %%
%%=============================================%%

%%=============================================================%%
%% Sample for another appendix section			       %%
%%=============================================================%%

%% \section{Example of another appendix section}\label{secA2}%
%% Appendices may be used for helpful, supporting or essential material that would otherwise 
%% clutter, break up or be distracting to the text. Appendices can consist of sections, figures, 
%% tables and equations etc.

\end{appendices}

\end{document}